%% file: main.tex
\ifdefined\pdfminorversion\pdfminorversion=7\fi
\documentclass{longharness_fairmeta}

\usepackage{longharness_preprint}

\usepackage{amsmath}
\usepackage{amssymb}
\usepackage{booktabs}
\usepackage{tabularx}
\usepackage{array}
\usepackage{enumitem}
\usepackage{algorithm}
\usepackage{algpseudocode}
\usepackage{xcolor}
\usepackage{colortbl}
\usepackage{multirow}
\usepackage{graphicx}
\usepackage{hyperref}
\usepackage{url}
\usepackage{listings}
\usepackage[most]{tcolorbox}

\newcommand{\method}{{\rmfamily\textsc{LongHarness}}}
\newcommand{\evidencebank}{{\rmfamily\textsc{EvidenceBank}}}
\newcommand{\conclusionstate}{{\rmfamily\textsc{Conclusion State}}}

\title{LongHarness: Advancing Agentic Reasoning over\\
100M-Token Corpora via Stateful Evidence Management}

\input{authors}

\abstract{
Corpus-level long-context reasoning requires agents to acquire and integrate
evidence across document collections that exceed their context windows.
Existing agent harnesses enable incremental access to these corpora, but
acquired evidence and evolving conclusions often remain entangled in a growing
interaction history.
This implicit state management can lead to two coupled failure modes:
unbounded exploration, where retrieval expands the interaction history
without resolving decision-relevant information gaps, and evidence-state
drift, where previously acquired evidence or subsequent corrections fail
to inform the current conclusion. We present \method{}, a
model-agnostic harness that makes the interaction between acquired evidence
and evolving conclusions explicit. Its core mechanism,
\emph{Evidence--Conclusion Coevolution}, uses unresolved conclusions to guide
exploration toward missing information, while newly acquired evidence
strengthens or revises the conclusions that directed the search.
\method{} couples this process with \emph{Evidence Lifecycle Management},
which keeps accumulated evidence current and traceable throughout the
interaction. Together, these mechanisms focus retrieval on remaining
information needs and keep conclusions aligned with current evidence. We
evaluate \method{} on OfficeQA Pro V1 and V2, AA-LCR, and LongBench v2
using GLM-5.3-Flash and Qwen3.8-Flash. Compared with direct reasoning using
the same backbone, \method{} improves GLM-5.3-Flash exact match on
OfficeQA Pro from 15.8\% to 60.2\% on V1 and from 4.4\% to 61.1\% on V2.
Against state-of-the-art agent harnesses, it surpasses Claude Code by
12.0 percentage points on V1 and Codex by 18.9 points on V2.
With Qwen3.8-Flash, \method{} reaches 61.7\% on V1, outperforming Claude
Code by 8.3 points and Codex by 12.0 points.
}

\begin{document}
\raggedbottom
\setlength{\textfloatsep}{18pt plus 3pt minus 3pt}

\hypersetup{pdftitle={LongHarness: Advancing Agentic Reasoning over 100M-Token Corpora via Stateful Evidence Management}}

\maketitle

\begin{figure}[H]
\centering
\includegraphics[width=\columnwidth]{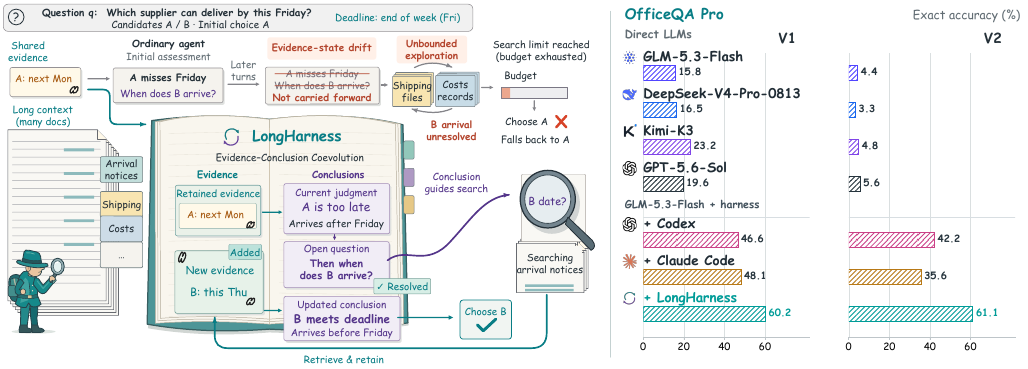}
\caption{Evidence lifecycle management for corpus-level long-context reasoning.
The left panel illustrates how an agent loses an earlier constraint and
continues searching, while \method{} retains the evidence, uses an
unresolved conclusion to guide retrieval, and revises its decision when
new evidence arrives. The right panel compares exact accuracy (\%) on
OfficeQA Pro V1 and V2 for direct models and agent harnesses using
GLM-5.3-Flash.}
\label{fig:loop}
\end{figure}
\clearpage

\input{sections/introduction}
\input{sections/method}
\input{sections/experiments}
\input{sections/related_work}
\input{sections/conclusion}

\bibliography{references}
\bibliographystyle{assets/plainnat}

\clearpage
\appendix
\input{sections/appendix}

\end{document}

%% file: authors.tex
\author[1,2]{\mbox{Songtao Huang}}
\author[2,4]{\mbox{Zhijie Zhong}}
\author[1,2]{\mbox{Xiaohan He}}
\author[1,2]{\mbox{Shuaiyu Zhang}}
\author[2,3]{\mbox{Zhiyin Yu}}
\author[2,4]{\mbox{Li Kang}}
\author[2,5]{\mbox{Heng Zhou}}
\author[2,6]{\mbox{Yutao Fan}}
\author[2]{\mbox{Xiangchao Yan}}
\author[2]{\mbox{Bo Zhang}}
\author[2,*]{\mbox{Shiyang Feng}}

\renewcommand\affiliationformat[2][]{{\small $^{#1}$#2}}
\renewcommand\affiliation[2][]{%
  \addtolist[#1]{#2}{\affiliationlist}{\affiliationformat}{\quad}%
}
\affiliation[1]{Fudan University}
\affiliation[2]{Shanghai Artificial Intelligence Laboratory}
\affiliation[3]{Peking University}
\affiliation[4]{Shanghai Jiaotong University}
\affiliation[5]{University of Science and Technology of China}
\affiliation[6]{Harbin Institute of Technology}

\contribution[*]{Corresponding author}
\metadata[Contact]{\email{huangsongtao@pjlab.org.cn}\quad\email{fengshiyang@pjlab.org.cn}}

%% file: sections/introduction.tex
\section{Introduction}
\label{sec:introduction}

Real-world document questions often require evidence scattered across large,
evolving document collections. Corpus-level long-context reasoning requires
agents to locate and integrate this evidence even when the collection exceeds
their context windows. Longer windows expand the material available within a
single invocation
\citep{reid2024gemini15,yang2025qwen25m,shang2025longrope2},
but models remain sensitive to evidence position and distractors even when
the relevant content fits
\citep{liu2024lostmiddle,hsieh2024ruler,ling2025longreason,zhuang2025docpuzzle}.
Agents must therefore search selectively, read relevant passages, and
accumulate evidence across multiple interactions. As these interactions
unfold, the challenge extends from finding evidence to keeping it usable,
revising it when needed, and carrying it into the final decision.

We view this process at three levels. \emph{Evidence retrieval} (Level~1)
locates relevant passages while filtering distractors. \emph{Evidence editing}
(Level~2) selects, reconciles, and revises retrieved content.
\emph{Evidence transformation} (Level~3) turns the resulting evidence into an
answer. Training methods supervise retrieval trajectories and memory edits
\citep{lin2026longtracerl,chen2026longrlvr,wang2025loongrl,gao2026maven},
while agent harnesses enable adaptive document access and persistent storage
\citep{cao2026codingagent,chen2023memwalker,lee2024readagent,zhao2024longagent,packer2023memgpt,karten2026prime,yu2026recuris}.
These advances strengthen evidence acquisition and maintenance, yet the
transition from accumulated evidence to an evolving conclusion often remains
implicit. The conclusions that determine what to search for next and what
answer to produce remain embedded in session history, free-form notes, or
transient reasoning.

At corpus scale, this implicit transition can lead to two coupled failure
modes. (i) \emph{Unbounded exploration} occurs when retrieval continues
without resolving decision-relevant information gaps, potentially exhausting
the interaction budget. (ii) \emph{Evidence-state drift} occurs when
previously acquired evidence or later corrections become disconnected from
subsequent reasoning, leaving the final decision based on stale or incomplete
evidence. These failures can reinforce each other as drift obscures which
information needs have already been resolved, prompting further retrieval
that expands the history and makes relevant evidence harder to recover.

We introduce \method{}, a model-agnostic harness that maintains
source-linked evidence and evolving conclusions as coupled persistent
states. Unresolved conclusion gaps guide subsequent search and reading,
while accepted evidence informs revisions to the conclusions and their
remaining gaps. Its core mechanism, \emph{Evidence--Conclusion
Coevolution}, uses unresolved conclusions to guide exploration toward
missing information, while newly acquired evidence strengthens or revises
the conclusions that directed the search. By making the remaining
information needs explicit, this feedback directs retrieval toward gaps
that matter to the decision and helps curb unbounded exploration.
\method{} couples this loop with \emph{Evidence Lifecycle Management},
which keeps accumulated evidence current and traceable throughout the
interaction. The \evidencebank{} retains source-linked statements and
tracks which remain active throughout the evidence lifecycle, helping
reduce evidence-state drift by keeping the basis for subsequent reasoning
aligned with the latest findings.

We evaluate \method{} on OfficeQA Pro V1 and V2, AA-LCR, and the LongBench
v2 Long subset using GLM-5.3-Flash and Qwen3.8-Flash. OfficeQA Pro requires
reasoning over roughly 100M tokens in V1, while V2 doubles the corpus and
introduces more demanding compositional questions. Compared with direct
reasoning using the same backbone, \method{} raises GLM-5.3-Flash exact
match from 15.8\% to 60.2\% on V1 and from 4.4\% to 61.1\% on V2.
Against state-of-the-art agent harnesses, it surpasses Claude Code by
12.0 percentage points on V1 and Codex by 18.9 points on V2.
With Qwen3.8-Flash, it reaches 61.7\% on V1, outperforming Claude Code by
8.3 points and Codex by 12.0 points. Trajectory analysis with GLM-5.3-Flash
further connects these gains to evidence management, with V2 success rates
exceeding the stronger of Claude Code and Codex at each stage by 11.1 points
in retrieval, 2.7 in evidence editing, and 9.3 in transformation.
Our contributions are summarized as follows.

\begin{itemize}[leftmargin=*,itemsep=2pt,topsep=3pt]
    \item We formulate corpus-level long-context reasoning as evidence
    lifecycle management, viewing the construction and continual refinement
    of the evidence supporting a decision as the central reasoning process.
    This perspective unifies evidence retrieval, editing, and transformation
    through the evolving role of evidence in decision-making.

    \item We develop \method{}, a model-agnostic harness that operationalizes
    this formulation through \emph{Evidence--Conclusion Coevolution} and
    \emph{Evidence Lifecycle Management}. It maintains evidence and
    conclusions as coupled persistent runtime states, allowing unresolved
    conclusions to direct evidence acquisition and newly acquired evidence
    to revise those conclusions, while an \evidencebank{} explicitly
    controls the evidence lifecycle and preserves the sources of active
    evidence.

    \item We evaluate \method{} on corpus-level reasoning over roughly
    100M-token and doubled document collections, together with
    provided-context benchmarks, using two model backbones.
    With GLM-5.3-Flash, \method{} outperforms Claude Code by 12.0 percentage
    points on OfficeQA Pro V1 and Codex by 18.9 points on V2, while ablations
    and trajectory analyses reveal the contributions of conclusion-guided
    exploration and evidence lifecycle management.
\end{itemize}

%% file: sections/method.tex
\section{Method}
\label{sec:method}

\subsection{Overview}
\label{sec:overview}

Given a question $q$ and documents $D=\{d_1,\ldots,d_n\}$ outside the
prompt, \method{} incrementally acquires evidence through an external
environment $\mathcal{E}_D$ under a resource budget. Its state
$S_t=(E_t,C_t,H_t)$ comprises source-linked evidence, evidence-linked
conclusions and unresolved gaps, and prior reasoning, actions, and
observations. Given $q$, document map $M_D$, state $S_t$, and remaining
budget $B_t$, the policy selects an action and an optional conclusion
update, and the environment returns the next state and observation.
\begin{align}
    (a_t,\widehat C_t) &\sim
    \pi_\theta(\cdot\mid q,M_D,S_t,B_t),
    \label{eq:agent-policy}\\
    (x_{t+1},o_{t+1}) &\sim
    \mathcal{P}_D(\cdot,\cdot\mid x_t,a_t),
    \label{eq:environment-transition}
\end{align}
Here, $x_t$ denotes the environment state, including registered source
spans, evidence entries and their lifecycle status, and resource counters.
Its active evidence forms $E_t$ in the policy state $S_t$.
The action $a_t$ specifies search, bounded reading, \evidencebank{} operations,
computation, or final submission, and $\widehat C_t$ is the optional
conclusion update. The controller validates action structure, registered
source references, and budget constraints before exposing updates to
subsequent turns.

\begin{figure}[t]
\centering
\includegraphics[width=\columnwidth]{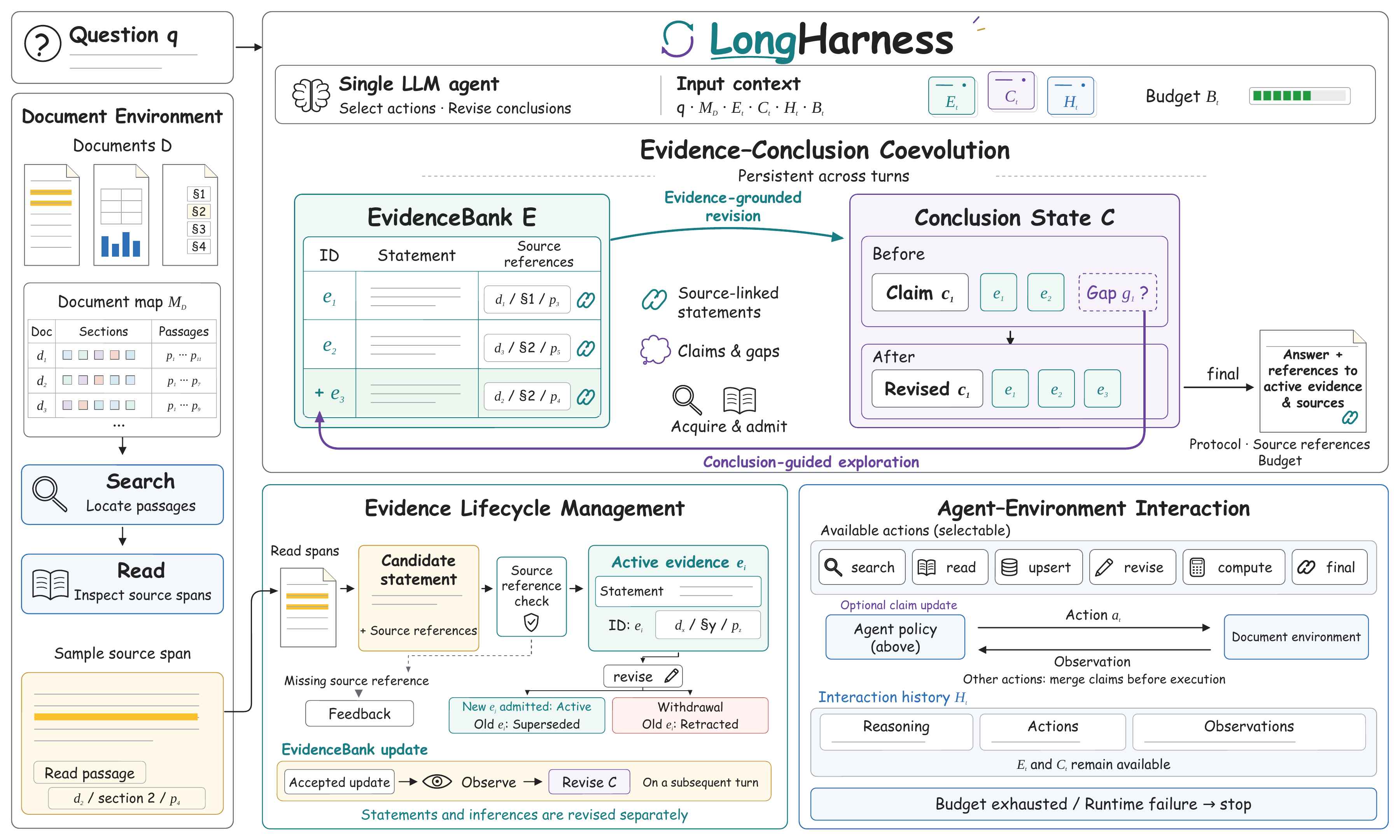}
\caption{Overview of \method{}. A single agent maintains an \evidencebank{} and
a \conclusionstate{} while interacting with external documents.
Evidence--Conclusion Coevolution uses unresolved conclusions to guide
exploration and accepted evidence to revise those conclusions, while
Evidence Lifecycle Management governs which statements remain active and
preserves their source references throughout the interaction.}
\label{fig:architecture}
\vspace{-9pt}
\end{figure}

\method{} couples two mechanisms (Figure~\ref{fig:architecture}).
Evidence--Conclusion Coevolution uses unresolved conclusions to guide
evidence acquisition and accepted evidence to strengthen or revise
conclusions (Section~\ref{sec:coevolution}). Evidence Lifecycle Management
controls evidence admission, revision, retraction, and supersession while
preserving source references (Section~\ref{sec:evidence-lifecycle}). A shared
interface supports adaptive interleaving of these operations with document
access, computation, and final submission
(Section~\ref{sec:agent-environment}).

\subsection{Evidence--Conclusion Coevolution}
\label{sec:coevolution}

\noindent\textbf{Separating evidence from inference.}
The \evidencebank{} maintains source-linked statements extracted from
passages the agent has read and serves as the persistent state in which
evidence is admitted and revised, while the \conclusionstate{} records the
agent's current interpretation of that evidence through established claims,
supported hypotheses, and unresolved gaps. Maintaining these representations
separately distinguishes the acquired evidence from the inferences drawn
from it, while explicit references connect each claim to its supporting
evidence. Each unresolved gap identifies information needed to establish,
qualify, or revise the current conclusion.
At the level of a reasoning cycle, the two states coevolve through
conclusion-guided evidence acquisition and evidence-grounded conclusion
revision,
\begin{align}
    E^{+} &= \mathcal{A}_\theta(q,D,E,C),
    \label{eq:conclusion-guided-exploration}\\
    C^{+} &= \mathcal{R}_\theta(q,E^{+},C).
    \label{eq:evidence-grounded-revision}
\end{align}
where $\mathcal{A}_\theta$ represents exploration guided by the current
conclusions and admission of the resulting statements into the
\evidencebank{}, while $\mathcal{R}_\theta$ updates the \conclusionstate{}
using the active evidence in the \evidencebank{}. The same agent policy
realizes both operations over one or more interactions, with the equations
specifying their dependencies while allowing exploration and revision to be
interleaved adaptively.

\noindent\textbf{Conclusion-guided exploration.}
The current \conclusionstate{} guides what the agent searches for and reads
next. For each claim, the agent identifies observations that would
establish, qualify, or refute it and uses these requirements to formulate
queries and select reading scopes. Competing hypotheses motivate searches
that distinguish between them, while unresolved gaps direct attention to
missing premises. When an observation leaves a gap unresolved, the agent
can reformulate the query or broaden the reading scope while keeping further
exploration tied to the information needed to advance the conclusion.

\noindent\textbf{Evidence-grounded revision.}
After observing the result of an \evidencebank{} update, the agent reassesses
the claims in the \conclusionstate{} against the bank's active evidence and
determines whether they remain supported, require qualification, or must be
revised. It updates the affected claims together with their evidence
references and unresolved gaps, keeping the \conclusionstate{} aligned with
the current evidence. Before final submission, the agent incorporates any
relevant evidence not yet reflected in this state and checks whether the
remaining gaps could change the requested decision.

\subsection{Evidence Lifecycle Management}
\label{sec:evidence-lifecycle}

Evidence Lifecycle Management governs how source-linked statements enter,
persist in, and leave the active evidence state in the \evidencebank{}.
A statement begins as a candidate extracted from a passage the agent has
read, becomes active upon admission, and may subsequently be revised or
retired as new evidence arrives. This lifecycle requires (i) traceability to
observed source passages, (ii) persistent availability across interactions,
and (iii) explicit updates when a statement's supporting evidence changes.

The \evidencebank{} implements these requirements through source-linked
admission, persistent storage, and explicit revision. For traceability,
each candidate statement must cite source references registered by prior
read actions, with the controller checking reference validity and the agent
judging whether the statement faithfully represents its sources. For
persistence, admitted statements remain available in the \evidencebank{}
across subsequent interactions, allowing the agent to reuse them without
reconstructing the evidence from earlier observations. For explicit
revision, a corrected statement is admitted as a new active entry and the
previous entry is marked as superseded, while a retracted statement leaves
the active state without a replacement. The agent observes these
\evidencebank{} updates before revising the affected conclusions, and final
references resolve to active statements and their underlying source passages.

\subsection{Agent--Environment Interaction}
\label{sec:agent-environment}

At each turn, the policy selects an action $a_t$ and an optional conclusion
update $\widehat C_t$ under the formulation in Section~\ref{sec:overview}.
The environment executes the action and returns an observation, while the
harness enforces the interaction protocol, source-reference constraints,
and resource limits.
The shared action space lets the policy interleave evidence acquisition,
maintenance, and use. \emph{Search} locates candidate documents or passages,
and \emph{Read} returns selected text and registers source references. The
agent proposes source-linked evidence through \texttt{upsert} and corrects
or retracts entries through \texttt{revise}
(Section~\ref{sec:evidence-lifecycle}). \emph{Compute} evaluates numerical
expressions supplied by the policy and validates their declared evidence
references, and \emph{Final} submits the answer with supporting references.

\evidencebank{} mutations defer conclusion updates until the agent observes
their results. The policy then revises affected claims and references through
optional partial updates, with omitted fields retained. Superseding or
retracting evidence does not automatically rewrite dependent conclusions.
The controller checks declared evidence references at computation and final
submission, while the policy remains responsible for source fidelity and
whether the evidence supports its conclusions. Reasoning, actions, and observations accumulate in
the history, while evidence and conclusions persist across turns. Final
acceptance requires a valid answer structure and references resolving to
active evidence and source passages. Resource exhaustion or runtime failure
may terminate the episode without an accepted answer.
Algorithm~\ref{alg:longharness} in Appendix~\ref{app:algorithm} summarizes
the procedure.

%% file: sections/experiments.tex
\section{Experiments}
\label{sec:experiments}
\label{sec:evaluation}

\suppressfloats[t]
\begin{table}[!t]
\setlength{\belowcaptionskip}{3pt}
\caption{Main results (\%) across four benchmarks. OfficeQA Pro reports exact
match (Ex.) and accuracy within 5\% relative error (@5\%), while AA-LCR
and the LongBench v2 Long subset report Exact accuracy. Direct baselines
use BM25 retrieval into a 256K-token window on OfficeQA Pro. ``-'' denotes
direct reasoning without a harness, and ``Avg.'' averages the Exact scores
across benchmarks. Each harness result is from one run per setting.
Direct AA-LCR scores are reported by Artificial Analysis; harness AA-LCR
scores are evaluated in this work.}
\label{tab:main-results}
\centering
\small
\definecolor{mainTableHeader}{HTML}{F1F4F7}
\renewcommand{\arraystretch}{1.18}
\setlength{\tabcolsep}{2.5pt}
\LongHarnessWideMainTable{
\toprule
\rowcolor{mainTableHeader}
\textbf{Model} & \textbf{Harness}
& \multicolumn{2}{c}{\textbf{OfficeQA V1}}
& \multicolumn{2}{c}{\textbf{OfficeQA V2}}
& \textbf{AA-LCR} & \textbf{LongBench V2} & \textbf{Avg.} \\
\rowcolor{mainTableHeader}
& & \textbf{Ex.} & \textbf{@5\%} & \textbf{Ex.} & \textbf{@5\%} & & & \\
\midrule
\mbox{Qwen3.8-27B} & - & 15.8 & 27.1 & 4.4 & 13.3 & 82.0 & 63.0 & 41.3 \\
\mbox{DeepSeek-V4-Flash-0731} & - & 14.3 & 27.1 & 3.3 & 6.7 & 79.7 & 53.7 & 37.8 \\
\mbox{Gemini-3.8-Flash} & - & 18.1 & 27.8 & 4.4 & 13.3 & 81.3 & 70.4 & 43.5 \\
\mbox{GPT-5.6-Luna} & - & 17.3 & 37.6 & 2.2 & 13.3 & 83.7 & 61.1 & 41.1 \\
\mbox{GLM-5.3} & - & 12.8 & 18.8 & 3.3 & 13.3 & 79.7 & 60.2 & 39.0 \\
\mbox{Qwen3.7-Max} & - & 13.5 & 27.8 & 3.3 & 22.2 & 79.0 & 64.8 & 40.2 \\
\mbox{Kimi-K3} & - & 23.2 & 34.7 & 4.8 & 16.1 & 88.7 & 68.0 & 46.2 \\
\mbox{DeepSeek-V4-Pro-0813} & - & 16.5 & 30.1 & 3.3 & 16.7 & 80.3 & 63.5 & 40.9 \\
\mbox{GPT-5.6-Sol} & - & 19.6 & 44.4 & 5.6 & 33.3 & 84.0 & 64.8 & 43.5 \\
\midrule
\mbox{GLM-5.3-Flash} & - & 15.8 & 27.8 & 4.4 & 15.6 & 80.0 & 62.0 & 40.6 \\
\mbox{GLM-5.3-Flash} & Codex & 46.6 & 61.7 & 42.2 & 62.2 & 77.0 & 69.4 & 58.8 \\
\mbox{GLM-5.3-Flash} & Claude Code & 48.1 & 67.7 & 35.6 & 58.9 & 79.0 & 63.0 & 56.4 \\
\mbox{GLM-5.3-Flash} & RLM & 45.1 & 62.4 & 38.9 & 58.9 & 56.0 & 60.2 & 50.0 \\
\rowcolor{cyan!6}
\mbox{GLM-5.3-Flash} & \textbf{\method{}} & \textbf{60.2} & \textbf{76.7} & \textbf{61.1} & \textbf{74.4} & \textbf{84.0} & \textbf{75.9} & \textbf{70.3} \\
\midrule
\mbox{Qwen3.8-Flash} & - & 15.8 & 28.6 & 4.4 & 15.6 & 80.0 & 54.3 & 38.6 \\
\mbox{Qwen3.8-Flash} & Codex & 49.6 & 69.2 & 50.0 & 70.0 & 81.0 & 69.4 & 62.5 \\
\mbox{Qwen3.8-Flash} & Claude Code & 53.4 & 72.9 & 41.1 & 52.2 & 83.0 & 63.9 & 60.3 \\
\mbox{Qwen3.8-Flash} & RLM & 48.1 & 65.4 & 35.6 & 48.9 & 73.0 & 69.4 & 56.5 \\
\rowcolor{cyan!6}
\mbox{Qwen3.8-Flash} & \textbf{\method{}} & \textbf{61.7} & \textbf{75.9} & \textbf{51.1} & \textbf{71.1} & \textbf{87.0} & \textbf{76.9} & \textbf{69.2} \\
\bottomrule
}
\end{table}

\subsection{Experimental Setup}
\label{sec:experimental-setup}

\paragraph{Benchmarks.}
We evaluate on OfficeQA Pro V1 and V2~\citep{opsahlong2026officeqapro},
AA-LCR~\citep{artificialanalysis2025aalcr}, and the Long subset of LongBench
v2~\citep{bai2024longbenchv2}, covering both corpus-level reasoning and
reasoning over provided contexts. OfficeQA Pro V1 contains 697 Treasury
documents totaling approximately 100M tokens, while V2 expands the corpus
to 1{,}435 documents and introduces composite statistical questions that
require combining values from multiple sources. Agents receive access to
the full corpus without question-specific gold documents, making source
discovery part of the task. AA-LCR provides the relevant context for each
of its 100 questions, while the LongBench v2 Long subset evaluates reasoning
over the benchmark's longest contexts. We report exact match and accuracy
within 5\% relative error on OfficeQA Pro, and Exact accuracy on AA-LCR and
LongBench v2.

\paragraph{Models and harnesses.}
Our direct-reasoning baselines include Qwen3.8-27B, Qwen3.7-Max, and
Qwen3.8-Flash~\citep{yang2025qwen3report}; DeepSeek-V4-Flash-0731 and
DeepSeek-V4-Pro-0813~\citep{deepseek2026v4};
Gemini-3.8-Flash~\citep{deepmind2026gemini38flash}; GPT-5.6-Luna and
GPT-5.6-Sol~\citep{singh2026gpt5card}; GLM-5.3 and
GLM-5.3-Flash~\citep{glm2026glm5}; and Kimi-K3~\citep{kimi2026k3}.
Citations refer to the corresponding model families or release-specific
model cards. On OfficeQA Pro, these models receive question-relevant
passages retrieved with BM25 into a 256K-token context window, while
LongBench v2 contexts follow the official truncation procedure with the
same token limit. We compare Codex~\citep{openai2026codexcli},
Claude Code~\citep{anthropic2026claudecode}, RLM~\citep{zhang2025rlm},
and \method{} using GLM-5.3-Flash and Qwen3.8-Flash as matched backbones,
holding model identity fixed within each comparison. All harnesses receive
the same questions and document corpus without access to reference answers
or gold source annotations.

\paragraph{Implementation details.}
All backbone models use temperature 1, top-p 0.95, maximum reasoning effort,
and an output limit of 32{,}768 tokens per request. Codex and Claude Code
each allow up to 200 model turns and 200 tool calls, while RLM allows
30 iterations and \method{} allows 200 steps subject to a cumulative read
budget of 1{,}000{,}000 tokens and a limit of 60 search calls. We grade
AA-LCR harness responses with GPT-5.6-Luna. Due to computational resource
constraints, the main comparison uses one run per harness and evaluation
setting. To assess variability across runs,
Appendix~\ref{app:repeated_runs} reports three-run results with
GLM-5.3-Flash, including individual scores, means, standard deviations,
and maxima. Missing answers and agent termination failures are counted
as incorrect.

\subsection{Main Results}
\label{sec:results}

Table~\ref{tab:main-results} shows that direct baselines reach at most
23.2\% exact match on OfficeQA Pro V1 and 5.6\% on V2 despite receiving
question-relevant passages in a 256K-token window. At corpus scale, this
fixed window cannot accommodate all potentially relevant evidence, and
passages selected before reasoning may omit information needed to resolve
intermediate conclusions. \method{} uses these conclusions to guide
further retrieval and accumulates the resulting evidence across
interactions, raising GLM-5.3-Flash exact match from 15.8\% to 60.2\% on
V1 and from 4.4\% to 61.1\% on V2.

With GLM-5.3-Flash, \method{} outperforms the strongest competing harness
by 12.0 percentage points on OfficeQA Pro V1 and 18.9 points on V2.
The larger advantage on V2 is consistent with maintaining
evidence across interactions as more scattered values must be located,
reconciled, and combined. With Qwen3.8-Flash, \method{} also leads on
both versions, reaching 61.7\% on V1 and 51.1\% on V2. Its advantage over
the strongest competing harness is 8.3 points on V1 but narrows to
1.1 points over Codex on V2, showing that the size of the gain varies
with the backbone and evaluation setting.
On the LongBench v2 Long subset, \method{} reaches 75.9\% with
GLM-5.3-Flash and 76.9\% with Qwen3.8-Flash, exceeding the next-best
harness by 6.5 and 7.4 percentage points, respectively. On AA-LCR, it
scores 84.0\% and 87.0\%, leading the strongest competing harness by
5.0 and 4.0 points. \method{} achieves the highest average Exact score
across the four evaluation settings with both backbones, reaching
70.3\% and 69.2\%, respectively.

\begin{table}[t]
\vspace{-6pt}
\begingroup
\setlength{\belowcaptionskip}{3pt}
\caption{Ablation results with GLM-5.3-Flash on OfficeQA Pro V1 and V2.
Ex. denotes exact-match accuracy (\%), and $\Delta$ denotes the decrease
from full \method{} in percentage points, computed before rounding.}
\label{tab:ablations}
\centering
\fontsize{9}{10.8}\selectfont
\renewcommand{\arraystretch}{1.22}
\setlength{\tabcolsep}{4pt}
\begin{tabular}{@{}l>{\hspace{4pt}}rr>{\hspace{10pt}}rr@{}}
\toprule
 \multirow{2}{*}{\textbf{Variant}}
 & \multicolumn{2}{c}{\textbf{OfficeQA Pro V1}}
 & \multicolumn{2}{c}{\textbf{OfficeQA Pro V2}} \\
\cmidrule(lr){2-3}\cmidrule(lr){4-5}
& \textbf{Ex.} & \textcolor{black!60}{$\Delta$}
& \textbf{Ex.} & \textcolor{black!60}{$\Delta$} \\
\midrule
\rowcolor{cyan!4}
\textbf{\method{} (full)} & \textbf{60.2} & \textcolor{black!60}{--}
& \textbf{61.1} & \textcolor{black!60}{--} \\
\addlinespace[3pt]
w/o Persistent States & 55.6 & \textcolor{black!60}{4.5}
& 53.3 & \textcolor{black!60}{7.8} \\
w/o Evidence--Conclusion Coevolution & 54.9 & \textcolor{black!60}{5.3}
& 51.7 & \textcolor{black!60}{9.4} \\
Naive Evidence Accumulation & 56.4 & \textcolor{black!60}{3.8}
& 51.1 & \textcolor{black!60}{10.0} \\
w/o Explicit Conclusion-Guided Exploration
& 55.6 & \textcolor{black!60}{4.5}
& 53.3 & \textcolor{black!60}{7.8} \\
Generic Working Notes & 53.4 & \textcolor{black!60}{6.8}
& 53.3 & \textcolor{black!60}{7.8} \\
\bottomrule
\end{tabular}
\par\endgroup
\end{table}

\subsection{Ablation Study}
\label{sec:ablations}

We ask whether evidence--conclusion coevolution and evidence lifecycle
management are critical to the effectiveness of \method{}, and examine
this question through five ablations in Table~\ref{tab:ablations}.
\textbf{(i) w/o Persistent States} removes the \evidencebank{} and
\conclusionstate{} while retaining interaction history.
\textbf{(ii) w/o Evidence--Conclusion Coevolution} retains the
\evidencebank{} but removes the \conclusionstate{} and its retrieval
feedback. \textbf{(iii) Naive Evidence Accumulation} preserves conclusion
feedback but permits only append-only evidence updates.
\textbf{(iv) w/o Explicit Conclusion-Guided Exploration} retains both
states but removes the instructions to guide retrieval using conclusion
gaps. \textbf{(v) Generic Working Notes} retains the \evidencebank{} and
replaces structured conclusions with editable free-text notes that can
record conclusions, unresolved questions, and evidence references.

Removing both persistent states reduces exact match by 4.5 percentage
points on V1 and 7.8 on V2, showing that explicit evidence and conclusion
states contribute beyond access to the interaction history. Removing the
maintained \conclusionstate{} and its retrieval feedback reduces accuracy
by 5.3 and 9.4 points. Restricting the \evidencebank{} to append-only
accumulation yields 56.4\% on V1 and 51.1\% on V2, corresponding to
decreases of 3.8 and 10.0 points. These results support the value of
revising and retiring evidence as new observations arrive, even when
conclusion feedback is preserved.
Removing the explicit conclusion-guided exploration instructions while
retaining both persistent states reduces accuracy by 4.5 and 7.8 points,
supporting the value of instructing the policy to use conclusion gaps to
guide retrieval beyond making those states available. Replacing structured
conclusions with generic working notes reduces accuracy by 6.8 and 7.8
points, showing that the full design outperforms editable free-text notes
that can also retain conclusions, open questions, and evidence references.
All five ablations incur larger losses on V2, consistent with the greater
importance of maintaining and reconciling evidence when questions require
combining values from multiple sources.

\pagebreak[3]
\begin{table}[t]
\vspace{-6pt}
\begingroup
\setlength{\belowcaptionskip}{3pt}
\caption{Success rates (\%) for evidence retrieval, editing, and transformation
on OfficeQA Pro V1 and V2 with GLM-5.3-Flash. Retrieval measures whether the
required evidence is found, editing measures whether retrieved evidence is
retained and adopted in the final decision, and transformation measures
whether the adopted evidence yields an exactly correct answer. Editing and
transformation rates are calculated only over cases that pass the preceding
stages. Each harness is evaluated on one run per split.}
\label{tab:evidence_pipeline}
\centering
\fontsize{9}{10.8}\selectfont
\renewcommand{\arraystretch}{1.18}
\setlength{\tabcolsep}{2pt}
\begin{tabular}{@{}lcccccc@{}}
\toprule
\multirow{2}{*}{\textbf{Evidence level}}
 & \multicolumn{3}{c}{\textbf{OfficeQA Pro V1}}
 & \multicolumn{3}{c}{\textbf{OfficeQA Pro V2}} \\
\cmidrule(lr){2-4}\cmidrule(lr){5-7}
 & Claude Code & Codex & \method{}
 & Claude Code & Codex & \method{} \\
\midrule
L1: Retrieval ($r_1$)
 & \textbf{85.0} & 72.2 & \textbf{85.0}
 & 72.2 & 76.7 & \textbf{87.8} \\
L2: Evidence editing ($r_2$)
 & 85.8 & 83.3 & \textbf{87.6}
 & 84.6 & 78.3 & \textbf{87.3} \\
L3: Transformation ($r_3$)
 & 66.0 & 77.5 & \textbf{80.8}
 & 58.2 & 70.4 & \textbf{79.7} \\
\bottomrule
\end{tabular}
\par\endgroup
\end{table}

\subsection{Interaction Length and Trajectory Analysis}
\label{sec:interaction}

We ask where agent harnesses fail across evidence retrieval, editing, and
transformation, and how these failures change from OfficeQA Pro V1 to V2.
For each harness, we use GPT-5.6-Luna at temperature 0 to audit incorrect
episodes and assess evidence retrieval, evidence adoption, and answer
transformation. Exact-correct episodes are counted as passing the first
two stages.
Let $E_1$ denote the presence of required evidence in the trajectory, $E_2$
its retention and reconciliation for use in the final decision, and $E_3$
an exactly correct answer. We measure retrieval coverage and conditional
editing and transformation success as
\begin{equation}
\label{eq:evidence_pipeline}
 r_1=P(E_1),\qquad
 r_2=P(E_2\mid E_1),\qquad
 r_3=P(E_3\mid E_1,E_2).
\end{equation}

\begin{figure}[!t]
\centering
\includegraphics[width=\linewidth]{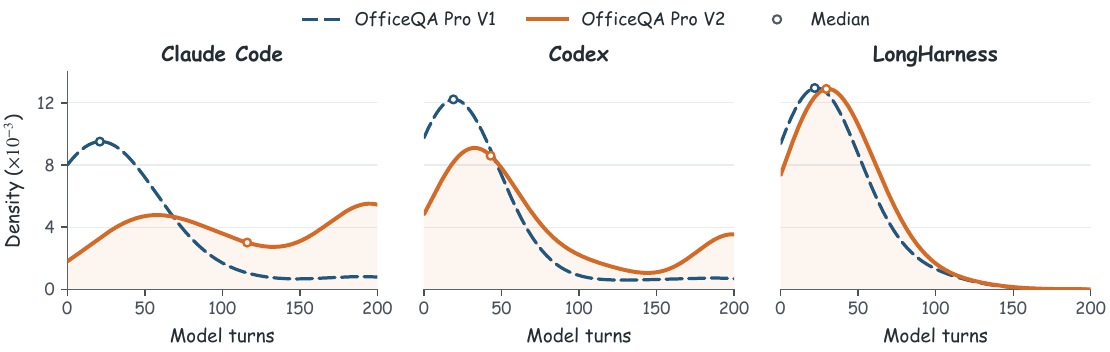}
\caption{Distributions of model turns per question on OfficeQA Pro V1 and V2
with GLM-5.3-Flash. Each panel compares the two versions for one harness.
Curves show kernel density estimates, and circles indicate medians.}
\label{fig:steps}
\par\vspace{12pt}
\centering
\includegraphics[width=\linewidth]{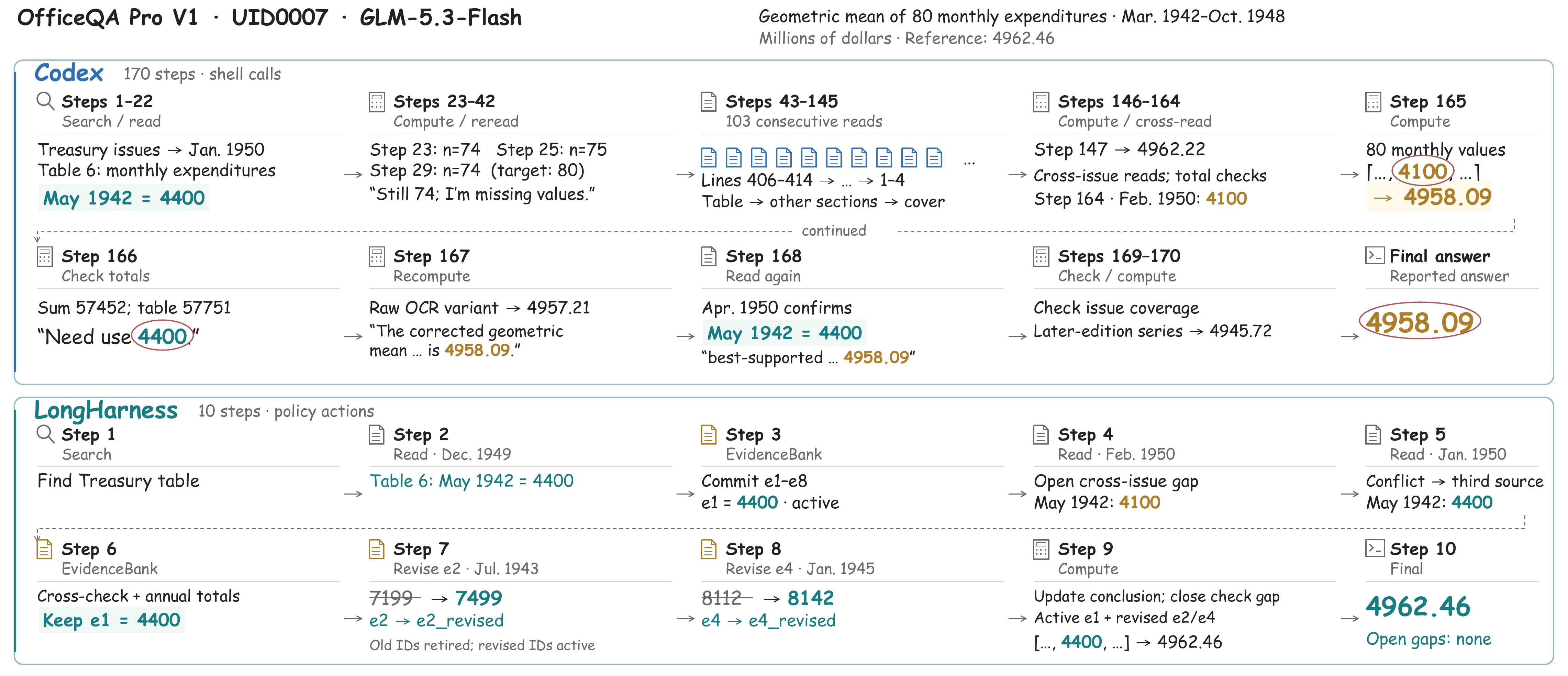}
\caption{Evidence lifecycle comparison with GLM-5.3-Flash on OfficeQA Pro V1.
For the geometric mean of 80 monthly Treasury expenditures, Codex continues
searching and omits a recognized correction; \method{} updates its evidence
and returns the reference answer, 4962.46 million dollars. Circles mark
Codex's incorrect input, correction, and final answer.}
\label{fig:case_study}
\end{figure}

Table~\ref{tab:evidence_pipeline} shows that \method{} matches Claude
Code's retrieval coverage on V1 and achieves higher conditional
evidence-adoption and answer-conversion rates. On V2, its retrieval,
evidence-adoption, and answer-conversion rates are 87.8\%, 87.3\%, and
79.7\%, respectively. The conditional adoption and conversion rates
remain close to their V1 values of 87.6\% and 80.8\%, indicating similar
evidence-use outcomes across the two evaluated settings. The conditional
rates are computed over the questions that pass the preceding stages for
each harness and describe stage outcomes within the analyzed runs.

Figure~\ref{fig:steps} shows that median turns increase from 22 to 30
for \method{}, from 21 to 116 for Claude Code, and from 19 to 43 for
Codex between V1 and V2. Codex exhausts its recorded model-turn budget
on 3.8\% of V1 questions and 21.1\% of V2 questions.
Claude Code's distribution also concentrates more heavily near 200 turns
on V2. V2 combines a larger corpus with questions requiring values from
multiple sources. \method{} records unresolved information needs in the
\conclusionstate{} and retains acquired evidence in the \evidencebank{},
allowing subsequent retrieval to focus on missing information while
reusing values already established. As questions require more cross-source
checks, this organization helps avoid repeatedly searching for or
re-establishing the same facts. This offers a plausible explanation for
the smaller shift in \method{}'s turn distribution from V1 to V2.
Figure~\ref{fig:case_study} compares how Codex and \method{} carry evidence
corrections into the final computation, with a detailed case analysis in
Appendix~\ref{sec:case_study}.

%% file: sections/related_work.tex
\section{Related Work}
\label{sec:related}

\subsection{Long-Context Reasoning}
\label{sec:related-lcr}

Post-training supervises evidence acquisition and use. LongRLVR rewards
correct identification of evidence chunks \citep{chen2026longrlvr}, LongPAS assigns advantages
to reasoning steps \citep{peng2026longpas}, and LongR measures document
consultation through information gain \citep{ping2026longr}. Training data
include key-fact chains \citep{wang2025loongrl}, distractor-augmented search
trajectories \citep{lin2026longtracerl}, capability-aligned task mixtures
\citep{lv2026golongrl}, and verifiable multi-hop or memory-management tasks
\citep{wan2025qwenlongl1,shen2025qwenlongl15}.
Training also uses agent-generated data and supervision for memory operations.
ACC compiles agent trajectories into long-context QA training examples
\citep{su2026acc}, AgenticLU distills self-generated clarification
trajectories without external teachers \citep{zhuang2025agenticlu},
MAVEN rewards evidence-memory edits \citep{gao2026maven}, and Memory-R2
credits individual memory operations \citep{yan2026memoryr2}. \method{}
instead couples evidence and conclusions at runtime, linking retrieval,
editing, and transformation without additional policy training.

\subsection{Agentic Harnesses}
\label{sec:related-harness}

ReAct couples reasoning with action \citep{yao2023react}, recursive language
models process context under program control \citep{zhang2025rlm}, and
Prime Agent maintains persistent REPL state \citep{karten2026prime}.
Codex and Claude Code support long-context reasoning \citep{cao2026codingagent};
other harnesses develop bounded long-horizon decomposition, harness
maintenance, and persistent memory
\citep{zheng2026onedayagent,lin2026ahe,packer2023memgpt,yu2026recuris,zhang2026uma,pan2026contextpilot}.
MemWalker uses memory trees \citep{chen2023memwalker}, ReadAgent uses
gist representations \citep{lee2024readagent}, GraphReader uses chunk
graphs \citep{li2024graphreader}, and multi-agent systems coordinate readers
\citep{zhao2024longagent,zhang2024coa}. SLIDERS reconciles evidence under
an induced schema \citep{joshi2026sliders}, ReContext replays found-but-unused
evidence \citep{zhao2026recontext}, and TRACE evaluates tool-augmented
reasoning trajectories \citep{kim2026trace}. LedgerMind uses grounding checks and event-triggered repair to address
unsupported or inconsistent claims~\citep{du2026ledgermind}. However,
existing claims can remain grounded while the evidence needed to answer
the question is still incomplete. In contrast, \method{} focuses on
corpus-level long-context reasoning, maintaining source-linked evidence
alongside evolving conclusions and unresolved information needs. These
needs guide subsequent search and reading, while newly acquired evidence
informs revisions to conclusions and remaining gaps.

%% file: sections/conclusion.tex
\section{Conclusion}
\label{sec:conclusion}

We propose \method{}, a model-agnostic harness that unifies evidence
retrieval, editing, and transformation for corpus-level long-context
reasoning. Evidence--Conclusion Coevolution
couples exploration with conclusion revision, while Evidence Lifecycle
Management keeps evidence current and traceable. Evaluations on OfficeQA
Pro, AA-LCR, and LongBench v2 show gains with two backbones. With
GLM-5.3-Flash, \method{} exceeds Claude Code by 12.0 percentage points on
OfficeQA Pro V1 and Codex by 18.9 points on V2. Ablations support the complementary
roles of persistent states, conclusion-guided retrieval, and evidence
revision. Trajectory analysis shows similar conditional evidence-use
rates across V1 and V2 and a smaller increase in interaction length for
\method{} than for the competing harnesses.

%% file: sections/appendix.tex
\section{Benchmark Details}
\label{app:benchmarks}

Table~\ref{tab:benchmark-details} summarizes the four evaluation settings,
covering external-corpus retrieval in OfficeQA Pro and reasoning over
provided contexts in AA-LCR and LongBench v2.

\begin{table}[!t]
\caption{Benchmark sizes, evidence access, and evaluation metrics. The AA-LCR
document count covers all document sets, and LongBench v2 lengths are
measured in words from the preprocessed contexts used in our experiments.}
\label{tab:benchmark-details}
\centering
\fontsize{9}{10.8}\selectfont
\definecolor{benchmarkTableHeader}{HTML}{F1F4F7}
\renewcommand{\arraystretch}{1.22}
\setlength{\tabcolsep}{4pt}
\begin{tabular}{@{}lrl@{\hspace{8pt}}l@{}}
\toprule
\rowcolor{benchmarkTableHeader}
\textbf{Benchmark} & \textbf{\# Items} & \textbf{Evidence access} &
\textbf{Metric} \\
\midrule
OfficeQA Pro V1 & 133 & 697-doc full corpus & Ex., @5\% \\
OfficeQA Pro V2 & 90 & 1{,}435-doc full corpus & Ex., @5\% \\
AA-LCR-1.1 & 100 & 234 provided docs & Judged Ex. \\
LongBench v2 Long & 108 & 132K--1.38M-word context & Choice Ex. \\
\bottomrule
\end{tabular}
\end{table}

\paragraph{OfficeQA Pro.}
OfficeQA Pro V1~\citep{opsahlong2026officeqapro} contains 133 questions
over 697 Treasury documents totaling approximately 100M tokens, while V2
contains 90 questions over an expanded corpus of 1{,}435 documents. V2
places greater emphasis on composite numerical questions requiring values
from different series, versions, or documents. Both versions expose the
full corpus without question-specific gold documents, requiring the system
to identify the sources needed for each answer. We report exact match,
which requires the requested value and format, and accuracy within 5\%
relative error.

\paragraph{AA-LCR-1.1.}
We use AA-LCR-1.1~\citep{artificialanalysis2025aalcr},\footnote{\url{https://huggingface.co/datasets/ArtificialAnalysis/AA-LCR}}
which contains 100 questions associated with 30 document sets comprising
234 documents in total, with an average of 99{,}325 tokens per set.
The documents include company and industry reports, government
consultations, academic papers, legal and marketing materials, and survey
reports. Questions require synthesis, temporal reasoning, conditional
logic, or numerical computation across documents rather than direct
retrieval from a single passage. Each question supplies its relevant
document set, allowing the evaluation to focus on evidence use within the
provided context. For direct-reasoning models, we report the AA-LCR scores
published by Artificial Analysis in its official evaluation results.

\paragraph{LongBench v2.}
LongBench v2~\citep{bai2024longbenchv2} contains 503 four-option questions
covering single-document QA, multi-document QA, long in-context learning,
long-dialogue history understanding, code-repository understanding, and
long structured-data understanding. The full benchmark spans contexts of
8K--2M words, most below 128K words. We evaluate the official Long subset
of 108 questions, whose contexts in the release used here range from
approximately 132K to 1.38M words with a median of 204K words. This subset
contains 29 code-repository understanding questions, 26 long in-context
learning questions, 24 single-document QA questions, 23 multi-document QA
questions, and 6 structured-data understanding questions. Each question
provides a long context and four answer options, and performance is
measured by exact accuracy on the selected option.

\section{Method Implementation Details}
\label{app:implementation}

This appendix specifies the interaction interface, \evidencebank{}
operations, state contract, and resource settings used in the reported
\method{} evaluations.

\subsection{Interaction Interface}
\label{app:interaction-interface}

\begin{table}[!t]
\caption{\method{} interaction interface, showing the purpose and returned
observation of each operation.}
\label{tab:interaction-interface}
\centering
\fontsize{8.8}{10.4}\selectfont
\definecolor{interfaceTableHeader}{HTML}{F1F4F7}
\renewcommand{\arraystretch}{1.18}
\setlength{\tabcolsep}{4pt}
\begin{tabularx}{\columnwidth}{@{}lXX@{}}
\toprule
\rowcolor{interfaceTableHeader}
\textbf{Operation} & \textbf{Purpose} & \textbf{Returned observation} \\
\midrule
\texttt{search} & Locate documents or passages relevant to an information
need & Ranked candidate locations and navigation snippets \\
\texttt{read} & Inspect selected locations under a bounded read extent &
Source text and registered source references \\
\evidencebank{} & Commit or revise source-linked statements & Active
statements, revision status, and validation feedback \\
\texttt{compute} & Evaluate a deterministic numerical expression over
evidence values & The computed value and integrity warnings \\
\texttt{final} & Submit the answer with supporting references & Accepted
output with resolved references, or validation feedback \\
\bottomrule
\end{tabularx}
\end{table}

The interface distinguishes candidate discovery from source inspection,
with \texttt{search} returning candidate locations and \texttt{read}
returning source text together with registered references. Search snippets
guide subsequent reading, while statements proposed to the \evidencebank{}
must cite references registered by \texttt{read}.

The policy selects one operation per turn, and the controller validates its
structure, source references, and resource requirements before execution.
Invalid actions return an error observation without modifying the
persistent states.

\subsection{EvidenceBank Operations}
\label{app:evidence-bank-operations}

\begin{table}[!t]
\caption{\evidencebank{} operations for admitting, correcting, and retracting
source-linked statements.}
\label{tab:evidence-bank-operations}
\centering
\fontsize{8.8}{10.4}\selectfont
\definecolor{evidenceOpTableHeader}{HTML}{F1F4F7}
\renewcommand{\arraystretch}{1.2}
\setlength{\tabcolsep}{4.5pt}
\begin{tabularx}{\columnwidth}{@{}lXX@{}}
\toprule
\rowcolor{evidenceOpTableHeader}
\textbf{Operation} & \textbf{Input} & \textbf{Effect} \\
\midrule
\texttt{upsert} & One or more proposed source-linked statements & Accepts
each validated statement as an active \evidencebank{} entry \\
\texttt{revise} & An active evidence identifier and either a replacement
statement or a null replacement & Replaces or withdraws the identified
active entry \\
\bottomrule
\end{tabularx}
\end{table}

An \texttt{upsert} validates each proposed statement independently by
checking that its source references were registered by earlier
\texttt{read} actions, so a batch may be partially accepted. The controller
validates reference registration, while the policy judges whether the
statement faithfully represents its sources.

Accepted \evidencebank{} entries retain immutable identifiers, and
corrections are recorded through \texttt{revise} rather than in-place edits.
A non-null replacement creates a corrected active entry and marks the
previous entry as superseded, while a null replacement retracts the previous
entry without creating a replacement. Superseded and retracted entries remain in the revision record but cannot
be cited in compute or final requests. Existing conclusions may retain
their references until the policy updates them.

\subsection{State Contract}
\label{app:state-contract}

The \conclusionstate{} contains established claims and supported hypotheses
linked to active \evidencebank{} entries, together with unresolved gaps
describing information that could change those conclusions. This
representation separates the agent's current inferences from the
source-linked statements maintained in the \evidencebank{}.

When an action updates the \evidencebank{}, conclusion revision follows
observation of the update result so that claims reference accepted evidence.
For other actions, the proposed conclusion update is merged before
execution. The complete interaction history remains available alongside the
persistent evidence and conclusion states throughout the episode, and final
supporting references must resolve to active \evidencebank{} entries and
their underlying source text.
Conclusion updates replace the supplied lists of established claims,
hypotheses, or open gaps. Omitted lists remain unchanged. When evidence
becomes inactive, transition logs flag affected conclusion references, but
the runtime does not automatically delete or rewrite those claims or block
subsequent search and reading. Compute and final validate the evidence IDs
declared in their requests, rather than all references retained in the
\conclusionstate{}. Compute also does not establish that literal operands
match the cited source values.

\subsection{Runtime Settings}
\label{app:runtime-settings}

\begin{table}[!t]
\caption{Interaction and reading limits used in the \method{} evaluations.}
\label{tab:runtime-settings}
\centering
\fontsize{9}{10.8}\selectfont
\definecolor{settingsTableHeader}{HTML}{F1F4F7}
\renewcommand{\arraystretch}{1.22}
\setlength{\tabcolsep}{5pt}
\begin{tabular}{@{}lr@{}}
\toprule
\rowcolor{settingsTableHeader}
\textbf{Setting} & \textbf{Value} \\
\midrule
Maximum turns & 200 \\
Maximum search calls & 60 \\
Cumulative read budget & 1{,}000{,}000 tokens \\
Maximum tokens per read & 16{,}384 \\
\bottomrule
\end{tabular}
\end{table}

At each step, the policy receives the system instructions and task
specification, followed by the complete interaction history and the current
runtime state. The history includes all previous reasoning, actions, and
tool observations. Historical observations are retained in full rather than
replaced with summaries or placeholders. The runtime state presents the
\evidencebank{}, \conclusionstate{}, and remaining resource budgets. These
explicit states provide an organized record of acquired evidence and
unresolved information needs alongside the full interaction history. The
cumulative read budget counts only source text returned by the \texttt{read}
operation across the episode.

The cumulative reading budget limits source acquisition over the episode;
each model request is separately subject to the endpoint's context limit.
The client does not automatically truncate or compact the interaction
history. A request failure that persists after the model client's retry
handling, including rejection of an oversized request, is recorded as
\texttt{model\_request\_failed} and terminates the episode. Episodes that
terminate without an accepted final answer are scored as incorrect.

The same state and action contract applies across benchmarks, with OfficeQA
Pro exposing the full external corpus and AA-LCR and LongBench v2 exposing
each question's provided context as the available document collection.
Reference answers, gold source annotations, and question-specific document
labels are withheld from the agent. All backbone models use temperature 1,
top-$p$ 0.95, maximum reasoning effort, and an output limit of 32{,}768 tokens
per request.

\subsection{Interaction Algorithm}
\label{app:algorithm}

Algorithm~\ref{alg:longharness} summarizes the interaction process, with
$\operatorname{Merge}$ replacing each supplied conclusion-state list
and preserving omitted lists (Appendix~\ref{app:state-contract}). For
example, an update containing only \texttt{open} replaces the unresolved
gaps while retaining the existing \texttt{established} and
\texttt{hypotheses} lists; an explicitly empty list clears that field.

\begin{algorithm}[H]
\caption{\method{} with Evidence--Conclusion Coevolution}
\label{alg:longharness}
\small
\begin{algorithmic}[1]
\Require Question $q$, documents $D$, policy $\pi_\theta$, budget $B_0$
\Ensure Accepted final output, or $\bot$ if none is accepted
\State Initialize document environment $\mathcal{E}_D$, document map $M_D$,
       and state $x_0$
\State $E_0 \gets \emptyset$, $C_0 \gets \emptyset$,
       $H_0 \gets \emptyset$, $t \gets 0$
\While{$B_t$ permits another turn and the environment remains operational}
    \State $S_t \gets (E_t,C_t,H_t)$
    \State $(a_t,\widehat C_t) \sim
           \pi_\theta(\cdot\mid q,M_D,S_t,B_t)$
    \Statex \hspace{\algorithmicindent}\Comment{Select action and optional conclusion update}
    \State $C_{t+1} \gets C_t$
    \If{$a_t$ violates protocol, source-reference, or budget constraints}
        \State $x_{t+1} \gets x_t$
        \State $o_{t+1} \gets$ validation feedback
        \Statex \hspace{\algorithmicindent}\hspace{\algorithmicindent}\Comment{Preserve persistent states on invalid actions}
    \Else
        \If{$a_t$ is neither \texttt{upsert} nor \texttt{revise}}
            \State $C_{t+1} \gets \operatorname{Merge}(C_t,\widehat C_t)$
            \Statex \hspace{\algorithmicindent}\hspace{\algorithmicindent}\hspace{\algorithmicindent}\Comment{Merge conclusion updates before other actions}
        \EndIf
        \State $(x_{t+1},o_{t+1}) \sim
               \mathcal{P}_D(\cdot,\cdot\mid x_t,a_t)$
        \Statex \hspace{\algorithmicindent}\hspace{\algorithmicindent}\Comment{Evidence updates are observed before conclusion revision}
    \EndIf
    \State $E_{t+1} \gets \operatorname{ActiveEvidence}(x_{t+1})$
    \State Extend $H_t$ with the turn's reasoning, action, and observation
           to obtain $H_{t+1}$
    \State $B_{t+1} \gets$ remaining budget after the turn
    \If{$o_{t+1}$ confirms an accepted final submission}
        \State \Return the final output and resolved source references
    \EndIf
    \State $t \gets t+1$
\EndWhile
\State \Return $\bot$
\end{algorithmic}
\end{algorithm}

\section{Experimental Protocol Details}
\label{app:protocol}

\subsection{Benchmark Preparation}
\label{app:benchmark-preparation}

OfficeQA Pro exposes the complete V1 or V2 corpus as an external document
environment, with a document map containing identifiers, titles, and
section structure. Agents must select relevant sources from this corpus
without access to question-specific gold documents.
For OfficeQA Pro V1 and V2, we use the Parsed Documents provided by the
benchmark authors.

AA-LCR-1.1 retains the document set supplied with each question, including
its official grouping and document order. Direct-reasoning baselines
receive these documents in the prompt, while agent harnesses access the
same collection through search and bounded reading. For LongBench v2,
direct-reasoning baselines follow the official truncation procedure with a
256K-token limit, while harnesses access the original Long-subset context
through the external document interface.

Reference answers, gold evidence annotations, and question-specific
document labels are excluded from agent-facing inputs, and document
filtering, ranking, and truncation do not use answer information or the
final prediction.

\paragraph{Document structure and retrieval.}
\method{} constructs its document environment from the benchmark text,
preserving source character offsets for subsequent evidence references.
Existing section annotations are retained. Otherwise, sections are
identified from Markdown headings, with a single root section used for
documents without headings. Searchable blocks follow paragraph boundaries,
with oversized paragraphs split into non-overlapping blocks of at most
1,024 estimated tokens, preferably at newline boundaries. Token estimates
count word sequences and individual non-whitespace punctuation symbols.

The search interface supports lexical and regular-expression queries,
optionally restricted to specified documents or sections. Lexical queries
use BM25-style ranking over the text blocks. Search results provide source
locators and short snippets, which guide subsequent \texttt{read}
operations. Read operations return source text together with registered
source-span identifiers, allowing \evidencebank{} entries to refer back to
the corresponding document offsets.

\subsection{Harness Configuration}
\label{app:harness-configuration}

All harnesses operate in an isolated workspace containing the question,
available documents, and an output directory. Tool-level network access is
disabled, and file access is confined to the workspace, excluding reference
answers, scoring scripts, and external run metadata.

\begin{table}[!t]
\caption{Available operations and resource limits for direct reasoning and the
agent harnesses in the main comparison.}
\label{tab:harness-configuration}
\centering
\fontsize{8.8}{10.4}\selectfont
\definecolor{harnessTableHeader}{HTML}{F1F4F7}
\renewcommand{\arraystretch}{1.18}
\setlength{\tabcolsep}{4pt}
\begin{tabularx}{\columnwidth}{@{}lXX@{}}
\toprule
\rowcolor{harnessTableHeader}
\textbf{System} & \textbf{Available operations} & \textbf{Constraints} \\
\midrule
Context window & Direct prompt-based reasoning over the supplied or retrieved
context & No tool interaction; 256K-token context window \\
Claude Code & Workspace file search and reading, local shell commands, and
workspace file modification & 200 model turns and 200 tool calls \\
Codex & Workspace file inspection, local shell commands, and workspace file
modification & 200 model turns and 200 tool calls \\
RLM & Persistent Python REPL, programmatic document access, and model
subcalls & 30 root-level iterations \\
\method{} & \texttt{search}, bounded \texttt{read}, \evidencebank{}
\texttt{upsert} and \texttt{revise}, \texttt{compute}, and \texttt{final} &
200 turns, 60 search calls, a 1M-token cumulative read budget, and at most
16{,}384 tokens per read \\
\bottomrule
\end{tabularx}
\end{table}

We use Codex CLI version 0.135.0 and Claude Code version 2.1.216.
Claude Code and Codex inspect and modify workspace files through their
native tools and local shell commands. RLM accesses documents
programmatically through its Python REPL. \method{} accesses documents
through its defined search and reading operations and maintains evidence
and conclusion states through explicit update actions.
Table~\ref{tab:harness-configuration} summarizes these interfaces and
resource limits. Each comparison uses the same questions, available
documents, and backbone model, with the harness interface and execution
protocol determining how the model interacts with the corpus.
RLM operates in a local Python REPL whose variables persist across
iterations within an episode. The model can inspect corpus files, process
retrieved text programmatically, and invoke model subcalls through the
REPL. The task prompt specifies the question, corpus location, and required
answer format. Our benchmark adapter converts document paths to absolute
paths accessible from the REPL and adds the benchmark-specific output
contract. The root loop allows up to 30 iterations. The driver uses the
library's default maximum recursion depth of one, at which subcalls receive
plain model completions rather than launching further REPL loops. Root
and subcalls use the same backbone. If the returned answer does not satisfy
the required output format, the adapter makes one additional call to the
same model, instructed to copy the answer value verbatim into the required
wrapper.

\paragraph{Evidence-pipeline audit.}
We use GPT-5.6-Luna with temperature 0 to assess incorrect episodes using
the recorded reasoning and available tool outputs or evidence-state
records. The audit distinguishes failures to retrieve the required
evidence, failures to select or retain that evidence, and failures to
transform the adopted evidence into a correct answer. Answer correctness
is determined by the benchmark scorer at tolerance 0; correct episodes
are counted as passing the first two stages by protocol. Retrieval
coverage is computed over all evaluated questions, evidence adoption over
questions passing retrieval, and answer conversion over questions passing
both preceding stages. These conditional rates characterize stage
outcomes within the analyzed runs.

\subsection{LongHarness Prompt Template}
\label{app:prompt-template}

The template below presents the instructions used by \method{} in the
terminology of the paper. A task-specific block supplies the question,
answer requirements, document map, tool schemas, and resource budget, with
the available operations defined in Appendix~\ref{app:interaction-interface}.

\paragraph{Runtime request construction.}
The prompt template supplies the policy instructions. Each model request
additionally contains a task block specifying the question, answer
requirements, document map, and action schemas. This is followed by all
previous assistant responses and tool observations in chronological order,
then a current-state block containing the \evidencebank{},
\conclusionstate{}, and remaining budgets. The \conclusionstate{} is
serialized under the \texttt{state} field, with \texttt{established},
\texttt{hypotheses}, and \texttt{open} lists. Historical observations remain
available alongside these explicit states.

\begin{tcolorbox}[
  enhanced,
  breakable,
  colback=black!3,
  colbacktitle=black!8,
  colframe=black!25,
  boxrule=0.3pt,
  arc=1.2mm,
  left=1mm,
  right=1mm,
  top=1mm,
  bottom=1mm,
  title={LongHarness Prompt Template},
  fonttitle=\small\bfseries\color{black}
]
\lstset{
  basicstyle=\ttfamily\footnotesize,
  breaklines=true,
  breakatwhitespace=true,
  columns=fullflexible,
  keepspaces=true,
  frame=none,
  backgroundcolor=\color{black!3},
  xleftmargin=1mm,
  xrightmargin=1mm,
  aboveskip=0pt,
  belowskip=0pt
}
\begin{lstlisting}
You are a ReAct agent for evidence-grounded long-context question answering. Documents remain external to the prompt. Investigate them over multiple turns using search, read, compute, EvidenceBank, and final. You decide what the evidence means, what remains uncertain, and when the answer is sufficient.

Follow one revisable loop:

1. Select the unresolved gap whose resolution is most likely to change the answer.
2. Use search to locate relevant documents and passages. Search snippets are navigation aids, not citable evidence.
3. Use read to inspect source text and register source references. Only a source reference returned by read may support an EvidenceBank statement.
4. Save useful source facts as atomic, source-linked EvidenceBank statements. An EvidenceBank identifier becomes usable only after a successful EvidenceBank observation.
5. Only cite EvidenceBank identifiers that were active before the current turn. Evidence created by the current EvidenceBank action becomes citable on the next turn.
6. Revise the Conclusion State when new evidence supports, qualifies, contradicts, or replaces earlier reasoning.

The EvidenceBank records what the documents state, not whether an answer is correct. Keep statements atomic and source-faithful. Retry only rejected statements. If an accepted statement is invalid, use EvidenceBank revise: provide a corrected replacement, or use a null replacement to withdraw it.

For compute, place the source values in one deterministic expression and cite the active EvidenceBank identifiers that support those values. After compute, verify direction, unit, and precision. Express the derived result in the Conclusion State together with the EvidenceBank identifiers supporting every operand.

Maintain a compact Conclusion State with three revisable lists:

- established: conclusions justified by active EvidenceBank entries;
- supported hypotheses: plausible but unsettled interpretations supported by active EvidenceBank entries;
- unresolved: concrete missing information that could change the answer.

Established conclusions and supported hypotheses must cite active EvidenceBank entries. Unsupported possibilities belong under unresolved. State the inference, not merely disconnected source facts.

Search for the missing distinction, not only for the question topic. If retrieval is uninformative, change the query or reading strategy instead of repeating it. Read enough context to interpret the source correctly. Before adopting a conclusion, verify that its entities, relationships, scope, and requested output match the question. A derived or aggregate conclusion must cite every decision-critical premise and use compute when arithmetic is needed. Evidence relevance alone is not support.

Before final, reconstruct the requested answer only from current established conclusions and their sources. Revise stale conclusions. Check a plausible alternative only when it could change the answer. Continue only while resolving a remaining unresolved gap could change the result. Do not search merely because unrelated uncertainty remains.

Place the answer-bearing conclusion under established. Submit final with its active EvidenceBank identifiers. Follow the requested choice label, units, precision, ordering, and output format.

Invoke exactly one operation per turn. Conclusion State updates may be partial, and omitted lists retain their previous values. Evidence mutation turns defer Conclusion State updates until their observations have been seen. Use final as the only way to submit the answer.
\end{lstlisting}
\end{tcolorbox}

\clearpage
\section{Repeated-Run Results}
\label{app:repeated_runs}

Table~\ref{tab:glm_repeated_runs} reports three runs for each harness and
evaluation setting with GLM-5.3-Flash. Summary statistics are computed
from the three reported scores. Std. denotes the sample standard deviation, and Max. is the highest
score across the three runs.

\begin{table}[!t]
\caption{Three-run Exact accuracy with GLM-5.3-Flash. Run scores, means,
and maxima are percentages. Standard deviations are in percentage points.}
\label{tab:glm_repeated_runs}
\centering
\small
\definecolor{mainTableHeader}{HTML}{F1F4F7}
\renewcommand{\arraystretch}{1.18}
\setlength{\tabcolsep}{10pt}
\begin{tabular}{@{}>{\columncolor{white}[0pt][\tabcolsep]}lccccc>{\columncolor{white}[\tabcolsep][0pt]}c@{}}
\toprule
\rowcolor{mainTableHeader}
\textbf{Harness} & \textbf{Run 1} & \textbf{Run 2} & \textbf{Run 3} & \textbf{Mean} & \textbf{Std.} & \textbf{Max.} \\
\midrule
\rowcolor{mainTableHeader}
\multicolumn{7}{c}{\textbf{OfficeQA Pro V1}} \\
\midrule
Codex & 45.86 & 46.62 & 47.37 & 46.62 & 0.76 & 47.37 \\
Claude Code & 48.12 & 48.87 & 51.13 & 49.37 & 1.57 & 51.13 \\
RLM & 45.11 & 42.86 & 48.12 & 45.36 & 2.64 & 48.12 \\
\rowcolor{cyan!6}
\method{} & 60.15 & 60.15 & 59.40 & 59.90 & 0.43 & 60.15 \\
\midrule
\rowcolor{mainTableHeader}
\multicolumn{7}{c}{\textbf{OfficeQA Pro V2}} \\
\midrule
Codex & 36.67 & 42.22 & 44.44 & 41.11 & 4.00 & 44.44 \\
Claude Code & 27.78 & 34.44 & 35.56 & 32.59 & 4.21 & 35.56 \\
RLM & 38.89 & 31.11 & 38.89 & 36.30 & 4.49 & 38.89 \\
\rowcolor{cyan!6}
\method{} & 61.11 & 58.89 & 58.89 & 59.63 & 1.28 & 61.11 \\
\midrule
\rowcolor{mainTableHeader}
\multicolumn{7}{c}{\textbf{AA-LCR 1.1}} \\
\midrule
Codex & 72.00 & 77.00 & 80.00 & 76.33 & 4.04 & 80.00 \\
Claude Code & 79.00 & 79.00 & 83.00 & 80.33 & 2.31 & 83.00 \\
RLM & 52.00 & 58.00 & 56.00 & 55.33 & 3.06 & 58.00 \\
\rowcolor{cyan!6}
\method{} & 86.00 & 84.00 & 82.00 & 84.00 & 2.00 & 86.00 \\
\midrule
\rowcolor{mainTableHeader}
\multicolumn{7}{c}{\textbf{LongBench v2 Long}} \\
\midrule
Codex & 68.52 & 69.44 & 69.44 & 69.13 & 0.53 & 69.44 \\
Claude Code & 62.04 & 62.96 & 64.81 & 63.27 & 1.41 & 64.81 \\
RLM & 59.26 & 60.19 & 62.96 & 60.80 & 1.92 & 62.96 \\
\rowcolor{cyan!6}
\method{} & 76.85 & 75.93 & 73.15 & 75.31 & 1.93 & 76.85 \\
\bottomrule
\end{tabular}
\end{table}

\section{Case Study}
\label{sec:case_study}

Figure~\ref{fig:case_study} compares GLM-5.3-Flash trajectories for computing
the geometric mean of 80 monthly Treasury expenditures in OfficeQA Pro V1.
We randomly selected this case from Codex trajectories that failed the
exact-match criterion. Steps in Figure~\ref{fig:case_study} denote shell-command
calls for Codex and policy actions for \method{}.
Codex locates the relevant table within 22 command calls but leaves missing
values implicit, later making 103 consecutive backward reads
(\emph{unbounded exploration}). It computes 4958.09 using 4100 for May 1942,
then recognizes the corrected value 4400 but retains the earlier result in
its final answer. Replacing this operand yields the reference answer
4962.46. Codex therefore fails exact match but passes the 5\% tolerance
criterion, with a relative error of approximately 0.088\%. This
\emph{evidence-state drift} illustrates a recognized correction that was
not incorporated into the final computation.

After recording 4400 in the \evidencebank{}, \method{} tracks cross-issue
consistency as an unresolved \conclusionstate{} gap. Further reading
corroborates the value and corrects two other entries. Evidence and
conclusions incorporate these revisions before computation. \method{}
returns 4962.46 in 10 policy actions, showing how conclusion-guided retrieval
and evidence maintenance carry corrections into the final decision.